\pdfoutput=1
\documentclass[10pt,twocolumn,letterpaper]{article}

\usepackage[pagenumbers]{wacv} % To force page numbers, e.g. for an arXiv version

\usepackage{natbib}
\usepackage[T1]{fontenc}
\definecolor{wacvblue}{rgb}{0.21,0.49,0.74}
\usepackage[pagebackref,breaklinks,colorlinks,allcolors=wacvblue]{hyperref}

\def\wacvPaperID{1772} % *** Enter the WACV Paper ID here
\def\confName{WACV}
\def\confYear{2026}

\title{Narrating For You: Prompt-guided Audio-visual Narrating Face Generation Employing Multi-entangled Latent Space}

\author{
    Aashish Chandra K\textsuperscript{1}\thanks{Equally contributing first authors.} \quad
    Aashutosh A V\textsuperscript{1,2}\footnotemark[1] \quad
    Abhijit Das\textsuperscript{1}\thanks{Corresponding author.} \\[0.5em] 
    \small
    \textsuperscript{1}Machine Intelligence Group, Department of CS\&IS, BITS Pilani, Hyderabad Campus, India\\
    \small
    \textsuperscript{2}Georgia Institute of Technology, USA \\
    {\tt\small \href{mailto:abhijit.das@hyderabad.bits-pilani.ac.in}{abhijit.das@hyderabad.bits-pilani.ac.in}}
}

\begin{document}
\maketitle
\begin{abstract}
\vspace{-4mm}

  We present a novel approach for generating realistic speaking and talking faces by synthesizing a person’s voice and facial movements from a static image, a voice profile, and a target text. The model encodes the prompt/driving text, the driving image, and the voice profile of an individual and then combines them to pass them to the multi-entangled latent space to foster key-value pairs and queries for the audio and video modality generation pipeline. The multi-entangled latent space is responsible for establishing the spatiotemporal person-specific features between the modalities. Further, entangled features are passed to the respective decoder of each modality for output audio and video generation. Our experiments and analysis through standard metrics demonstrate the effectiveness of our model. All model checkpoints, code, and the proposed dataset can be found at \url{https://github.com/narratingForYou/NarratingForYou}.
\end{abstract}

\section{Introduction}
\label{sec:intro}

AI-generated real-time audio-video multimedia communication by rendering realistic human talking faces has recently drawn massive attention \cite{1, 2}. Such technology is promising in various applications such as digital communication, aiding communication with individuals with impairments, designing artificial instructors, and developing interactive healthcare \citep{xu2024hallohierarchicalaudiodrivenvisual,gan2023efficientemotionaladaptationaudiodriven}. In such applications, generating realistic and real-time speech and visual content simultaneously is a key requirement.
%
% -- See Alternatives below
%
Therefore, in an ideal scenario, given a prompt text along with a face image and the audio profile of an individual, a talking human face would be rendered as output with audio (generated speech) and visual narration according to the prompt text.
%
%
% Ali ()
% Therefore, an ideal scenario would be that given a prompt text along with a face image and the audio profile of an individual, a talking human face is screen-played as output with visual narrating according to the prompt text.
%

\label{sec:intro}

%\begin{figure*}
%    \centering
    %\advance\leftskip-6cm
%    \includegraphics[width=14cm]{imgs/Fig1.png}
%    \caption{SOTA approaches of talking face generation use a face image as driving frame, with an audio prompt passed as input to the existing model such as Hallo \citep{xu2024hallohierarchicalaudiodrivenvisual}, VASA \citep{xu2024vasa1lifelikeaudiodriventalking} and the proposed model which generates a realistic audio-video synchronous multimodal talking face with face image and audio profile of an individual along with the prompt text.}
%    \label{fig:teaser}
%\end{figure*}

Generative AI has emerged as a key area of interest in the computer vision, multimedia, learning representation, and machine learning community. Although existing approaches have made significant strides, they are constrained by their reliance on generating a single modality \citep{egger20203dmorphablefacemodels,kim2021conditionalvariationalautoencoderadversarial}. For example, current text-to-speech models (TTSM) \citep{ao2022speecht5unifiedmodalencoderdecoderpretraining,betker2022tortoise,casanova2024xttsmassivelymultilingualzeroshot} focus primarily on voice synthesis. %, often neglecting the nuanced interplay between speech characteristics and the face of a person. 
Similarly, visual generation techniques i.e. talking face models (TFM) \citep{
ren2021pirenderercontrollableportraitimage,
rombach2022highresolutionimagesynthesislatent, siarohin2020ordermotionmodelimage, zhang2023dreamtalkdiffusionbasedrealisticemotional,xu2024hallohierarchicalaudiodrivenvisual, xu2024vasa1lifelikeaudiodriventalking,   zhang2023sadtalkerlearningrealistic3d} aim at face video generation given a text or/and audio or/and image as a prompt. Hence, both TTSM and TFM techniques are unsuitable for real-life audio-video multimedia communication scenarios such as audio-visual chatbots, as in such situations, both realistic video and speech must be generated synchronously and simultaneously. Few efforts have been made in the literature to merge TTSM and TFM by cascading the pipeline \citep{wang2023text, zhang2022text2video}. Additionally, \citep{jang2024faces} made an effort to generate a talking face and speaking audio jointly for a specific individual from a prompt text. However, as the audio generation is not based on an audio prompt it is not generalized for real-life applications.  

Further, these TFM \citep{chen2024anifacediffhighfidelityfacereenactment,zhang2019oneshotfacereenactment} depend on guidance from defined facial properties from the weakly supervised latent information from the reference modality. As a result, poor lip-synchronization and limited ability to tune an existing audio profile for personalizing the video content lead to a generation that is far from being realistic. Moreover, expressiveness in facial dynamics, along with subtle nuances for realistic facial behavior, needs to match simultaneously with audio content temporally to produce realistic talking faces. Further, such synchronization also depends on individual traits, such as speech intonation and other covariates. Although they are supposed to be important considerations for realistic speaking and taking faces models (STFM), However, this was not in the scope of existing work on STFM \citep{jang2024faces}. Therefore, this gap in the literature motivates us to design a prompt text-guided audio-visual multimodal generative STFM that can jointly generate audio and video, given a reference image and reference audio along with the prompt text as input. 

Consequently, in contrast to existing literature, in this work, we introduce a novel multi-modal framework designed to address these limitations by generating highly realistic speech and animations from a combination of prompt text, a driving image, and an audio profile as inputs. Specifically, our framework aims to synthesize videos of a talking human face where the person in the image appears to speak along with the generated voice from the provided text for the given identity. Our method enhances the capabilities of existing pre-trained models \citep{xu2024hallohierarchicalaudiodrivenvisual} with an advanced parallel mechanism that leverages both visual and auditory data streams. This parallelism ensures that the synthesized videos not only align the subject’s facial movements with the spoken text but also synchronize with the generated personalized voice outputs that correspond to the subject’s appearance.

A person-agnostic generalized STFM model must encompass a large appearance and acoustic features variation. Furthermore, extracting such structure information along with the temporal synergy between the audio and video while preserving individual variance requires additional modules to model these complexities. Therefore, we introduce a parallel multiple entanglement in the latent space between the encoding and decoding of different modalities. The encoding stage is designed to generate structured representations that are enhanced in the latent space to generate the target representation in the decoding stage.

Our proposed architecture for STFM contains three main phases (See Figure~\ref{fig:NetworkArchitecture}). \textit{Modality encoding phase}, at this stage a heterogeneous personal signature of the audio and video modality, and the driving feature from the text are extracted. The second stage is the \textit{multi-entangled latent space} which glens the spatiotemporal relation and synchronization in the embeddings of the modalities, which further acts as the input to the \textit{decoders phase} i.e the third stage of the proposed architecture. In the second stage, the exchange of information between the key and values (identity information from audio and video extracted from the individual encoders) and queries (driving features from encoded prompt text) are streamlined. To instrument this, an entanglement of the audio and text latent is performed, which further entangles with video latent in the transformer block and then to a diffusion block. The output of the diffusion block is passed to the video decoder. Similarly, an entanglement of the video and text latent is performed, which further entangles with audio latent in a transformer space and passes to a text decoder block and then to the audio decoder. Such entanglements ensure to streamlining of the audio profile and the driving image by linear navigation in the latent space along with the encoded feature from the prompt text. Specifically, the temporal information for both the audio and video generation is constructed by linear displacement of codes in the latent space as per the encoded text prompt. In turn, the model also learns a set of orthogonal motion directions to simultaneously learn the audio and video temporal synergy by exchanging their linear combination to represent any displacement in the latent space. 
%There are several dataset available in the literature with audio-video contents for taking faces such as VoxCeleb \citep{Nagrani19}, FakeAVCeleb \citep{khalid2022fakeavcelebnovelaudiovideomultimodal}, HDTF \citep{zhang2021flow} and CelebV-HQ \citep{zhu2022celebvhqlargescalevideofacial}. All these datasets do not in the text prompt. Hence, we developed a domain-specific dataset for the problem. 
To summarize, our key contributions are as follows:
\vspace{-2mm}
\begin{itemize}
    \item To the best of our knowledge, the proposed architecture is the first person-agnostic STFM which fosters a text-driven multimodal realistic audio-video synthesis that can be generalized to any identity.
    \item We design a three-phase architecture which consists of the encoder, multi-entangled latent and decoder phase for audio and video pipeline. The muti-entangled latent space glens the spatiotemporal and synchronisation in the encoder embedding to exchange information between the modality and guided text and help to generate crucial visual and acoustic characteristics based on input profiles.
    %\item We also produced a dataset specific to STFM scenario with varying audio and video quality and a wide variety of text prompts. 
    \item With the comprehensive experiments,  we demonstrate that the proposed method surpasses the state-of-the-art techniques available for STFM.
\end{itemize}

\vspace{-2mm}
\section{Related Work}
\vspace{-1mm}
%Text-to-speech (TTS)  technology has seen remarkable progress in recent years, with the development of models that generate highly natural and expressive speech. 
Modern Text-To-Speech approaches as\citep{casanova2024xttsmassivelymultilingualzeroshot,betker2022tortoise}, Tacitron\citep{wang2017tacotronendtoendspeechsynthesis} and the newer Tacitron2\citep{shen2018naturalttssynthesisconditioning} are quite popular. The spectrograms generated from these models are then passed through neural vocoders like WaveNet\citep{oord2016wavenetgenerativemodelraw} or HiFi-GAN\citep{kong2020hifigangenerativeadversarialnetworks} to generate high-quality audio waveforms. Other models, such as FastSpeech\citep{ren2019fastspeechfastrobustcontrollable} and VITS\citep{kim2021conditionalvariationalautoencoderadversarial}, are noteworthy for improving the speed of speech generation. TortoiseTTS\citep{betker2022tortoise} is a modern, expressive TTS system with impressive voice cloning capabilities which incorporates a combination of the Auto-Regressive Model, followed by a Diffusion Model\citep{ho2020denoisingdiffusionprobabilisticmodels}. This model also follows the standard of a vocoder(Univnet)\citep{jang2021univnetneuralvocodermultiresolution} for generating the audio from the spectrogram frames. Only a few works have been made in the literature to attend STFM by cascading the pipeline \citep{wang2023text, zhang2022text2video}. In \citep{jang2024faces} advancements are made by generating a talking face and speaking audio jointly for a specific individual from a prompt text.  
Face reenactment and lip-sync models, such as SyncNet\citep{raina2022syncnetusingcausalconvolutions}, focused on lip synchronisation through facial key points and phoneme mapping. More recent models, such as LipGAN\citep{K_R_2019} and Wav2Lip\citep{Prajwal_2020}, leverage GANs to improve lip-sync accuracy.
%The multimodal synthesis of human videos, combining text, audio, and visual inputs, has advanced considerably in recent years. Early approaches focused on audio-driven models that primarily addressed lip-syncing, mapping speech inputs to corresponding facial movements. Models like  SyncNet\citep{raina2022syncnetusingcausalconvolutions} played a crucial role in establishing baseline synchronization between audio and lip movements. However, these models often lacked expressive, natural face dynamics.
Diffusion-based lip-Sync models such as Audio2Head\citep{wang2021audio2headaudiodrivenoneshottalkinghead}, Expressive Audio-driven Talking-heads (EAT)\citep{gan2023efficientemotionaladaptationaudiodriven},  Hallo\citep{xu2024hallohierarchicalaudiodrivenvisual},  SadTalker\citep{zhang2023sadtalkerlearningrealistic3d}, FaceChain ImagineID\citep{xu2024facechainimagineidfreelycraftinghighfidelity},  Diffused Heads\citep{stypułkowski2023diffusedheadsdiffusionmodels} and DreamTalk\citep{zhang2023dreamtalkdiffusionbasedrealisticemotional} have been proposed in the literature of TFM. 

\section{Proposed Architecture for STFM}

\begin{figure*}[t]
\centering
\includegraphics[width=0.95\textwidth]{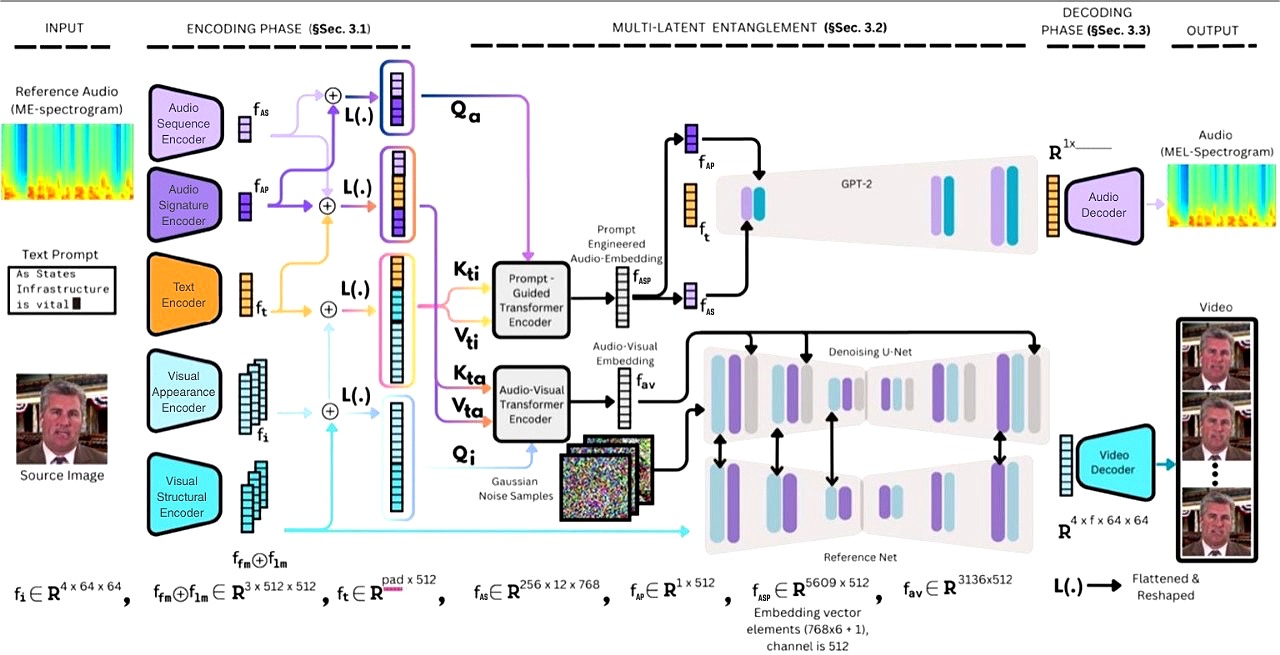}
\caption{\textbf{Our Network Architecture: } Text Prompt-guided joint audio-visual learning representations using dual stream Transformer Encoders and Denoising Diffusion model. The model architecture can be divided into three phases -- namely \textit{Encoding Phase}, \textit{Multi-Latent Entanglement}, and \textit{Decoding Phase}. As an output, an audio-visual animation is generated from a single source image, reference audio, and a short text prompt.}
\label{fig:NetworkArchitecture}
\end{figure*}

We propose an architecture for STFM for joint learning methodology for the audio, video, and natural language-based text prompts consisting of three main components, namely, \textbf{(1)} Encoding phase, \textbf{(2)} Entanglement of combined latent space, and \textbf{(3)} Decoding phase \textit{i.e.,} Latent conditional generation of synthesised audio-video. Figure \ref{fig:NetworkArchitecture} illustrates the detailed network architecture and roles of different model components to learn and dynamically synthesise audio and video on a given source image. 

\subsection{Multi-modal Encoding Phase. }\label{subsec:EncodingSpace}
The encoder stage aims to glean the heterogeneous personal signature and the coded latent form as per the length of the output sequence of the audio and video modality, along with the driving encoded feature from the text. 

Hence, to encode the audio modality, two sets of encoders are employed. One set will encode the personal signature of the modality for the respective individual whose source image and reference audio are passed as input, and the second set of encoders features the input as per the length of the output sequence. For the audio sequence encoding $\boldsymbol{\mathrm{E}_{AS}}$, we use the HiFi-GAN encoder \citep{kong2020hifigangenerativeadversarialnetworks}, which takes a mel-spectrogram as input and upsamples it through transposed convolutions to the length of the output sequence as per the temporal resolution of raw waveforms. For the audio signature/profile encoding $\boldsymbol{\mathrm{E}_{AP}}$, we employed Wav2Vec Encoder \citep{baevski2020wav2vec20frameworkselfsupervised} to extract high-dimensional heterogeneous personal embedding vectors from the reference audio. The $\boldsymbol{\mathrm{E}_{AS}}$ generates a feature embedding $\boldsymbol{\mathrm{f}_{AS}}$ that represents the output sequence audio waveform. At the same time, the $\boldsymbol{\mathrm{E}_{AP}}$ produces a personal audio signature embedding $\boldsymbol{\mathrm{f}_{AP}}$  capturing semantic of individual specific audio information. Hence, the semantic audio embedding is a direct mapping of the speaker's voice profile. Consequently, the combined features $\boldsymbol{\mathrm{f}_{AS}} \oplus \boldsymbol{\mathrm{f}_{AP}}$ provide a detailed feature $\boldsymbol{\mathrm{f}_{ASP}}$ as per the output sequence along with audio profile necessary for driving the audio generation.% the lip-sync and facial animations in the synthesized video. 
The input reference audio is represented as a 2-second MEL-spectrogram, encoder into a sequence of acoustic features per frame of 0.2 seconds duration with the shape of $\mathbb{R}^{5609 \times 512}$.  

For the input text prompt is encoding by $\boldsymbol{\mathrm{E}_{T}}$ employing a Byte-Pair Encoding (BPE) and Tokenization \citep{zouhar2024formalperspectivebytepairencoding} to convert textual information into a feature vector $\boldsymbol{\mathrm{f}_{t}} \in \mathbb{R}^{512.T}$. This feature vector enables output context-specific information, allowing the synthesized video and audio to align with the intended spoken words and expressions implied in the text. To enable the audio feature vector with the output context a concatenating $\boldsymbol{\mathrm{f}_{t}}$ with the combined feature of reference audio $\boldsymbol{\mathrm{f}_{AS}} \oplus \boldsymbol{\mathrm{f}_{AP}}$ is to obtain the speaker's signature in the final flattened feature tokens of $\boldsymbol{\mathrm{f}_{t}} \oplus 
\boldsymbol{\mathrm{f}_{AS}} \oplus \boldsymbol{\mathrm{f}_{AP}} \in \mathbb{R}^{5609+T \times 512}$.

For a video sequence, two encoders are employed: a visual appearance encoder $\boldsymbol{\mathrm{E}_{VA}}$ and the visual structural encoding $\boldsymbol{\mathrm{E}_{VS}}$ to process the input source image. For $\boldsymbol{\mathrm{E}_{VA}}$ a Variational Auto-Encoder (VAE) \citep{kingma2022autoencodingvariationalbayes} which encodes the appearance feature of the input and a landmarks detection model \citep{zhang2020mediapipehandsondevicerealtime} for $\boldsymbol{\mathrm{E}_{VS}}$ which can capture the structural feature of the profile. The VAE generates an image embedding $\boldsymbol{\mathrm{f}_{i}}$, representing the visual style and identity of the person in the source image. Concurrently, the landmarks detection network extracts structural features face mask feature $\boldsymbol{\mathrm{f}_{fm}}$ and lip mask feature $\boldsymbol{\mathrm{f}_{lm}}$, which are combined with the image embedding vectors to create a fused visual feature representation $\boldsymbol{\mathrm{f}_{i}} \oplus \boldsymbol{\mathrm{f}_{lm}} \oplus \boldsymbol{\mathrm{f}_{fm}} \in \mathbb{R}^{3136 \times 512}$. The straightforward tendency of traditional methods is either to introduce prior 3D morphable models faces \citep{zhang2023sadtalkerlearningrealistic3d}, motion priors of the facial parts \citep{jang2024faces}, or guiding video frames \citep{wang2022latent} to learn nuances of facial articulation concerning the audio in the combined latent space. In contrast, we show that the entanglement of multiple latent spaces of text-audio-video using Transformer encoders \citep{vaswani2023attentionneed} can eliminate the dependency on strong motion priors. As a result, we are able to use text prompt features as a set of anchoring tokens to both the Transformer encoders. To enable the visual feature vector with the output context, a concatenation $\boldsymbol{\mathrm{f}_{t}}$ with the combined feature of reference visual feature $\boldsymbol{\mathrm{f}_{i}} \oplus \boldsymbol{\mathrm{f}_{lm}} \oplus \boldsymbol{\mathrm{f}_{fm}}$.

\subsection{Entanglement of Multi-modal Latent Space }\label{subsec:EntagledLatentSpace} 
As illustrated in Figure~\ref{fig:NetworkArchitecture}, a smooth synergy between the text-audio latent embedding and the text-image latent embedding is established by two Transformer encoders followed by a latent diffusion-guided \citep{xu2024hallohierarchicalaudiodrivenvisual} synthesiser of visual nuances and decoder employing GPT-2 decoder \citep{casanova2024xttsmassivelymultilingualzeroshot} to model text-conditioned of the audio latent. 

The first Transformer encoder spatially contextualizes the audio MEL-spectrogram tokens using a dual-stream cross-modal attention (CA) mechanism with the flattened version, denoted by $\boldsymbol{\mathrm{L}(.)}$, of \textit{categorically fixed speaker} embedding tokens merged with varying text embedding tokens,\textit{ i.e.,} $\boldsymbol{\mathrm{Q}_a} = \boldsymbol{\mathrm{L}}(\boldsymbol{\mathrm{f}_{AS}} \oplus \boldsymbol{\mathrm{f}_{AP}})$, as 
\begin{equation}\label{eq:crossAttn1}
    \text{CA}(\boldsymbol{\mathrm{Q}_a}, \boldsymbol{\mathrm{K}_{ti}}, \boldsymbol{\mathrm{V}_{ti}}) = \text{SoftMax}\left(\frac{\boldsymbol{\mathrm{Q}_a}\boldsymbol{\mathrm{K}_{ti}}^\top}{\sqrt{d_k}}\right) \boldsymbol{\mathrm{V}_{ti}},
\end{equation}
where the query vector $\boldsymbol{\mathrm{Q}_a}$ is of dimension $\mathbb{R}^{5609\times512}$ and the key-value paring ($\boldsymbol{\mathrm{K}_{ti}}, \boldsymbol{\mathrm{V}_{ti}}$) between the tokens of $\boldsymbol{\mathrm{L}}( \boldsymbol{\mathrm{f}_{t}} \oplus 
\boldsymbol{\mathrm{f}_{i}} \oplus \boldsymbol{\mathrm{f}_{lm}} \oplus \boldsymbol{\mathrm{f}_{fm}})$ has a variable spatial length (padded up-to a max length) with a fixed channel length of 512. Merging the varying text tokens serves two purposes -- \textbf{(1)} first, querying audio tokens as well as the speaker tokens has been implicitly prompt-engineered by the text tokens, \textbf{(2)} second, when the resulting prompt-engineered latent embedding vectors $\boldsymbol{\mathrm{f}_{as}}$ are split into their respective constituents, they become proxy weights of text-image embedding vectors. 

Similar to the previous encoder block, the second Transformer encoder spatially contextualizes the input masked-image embedding vectors $\boldsymbol{\mathrm{L}}(\boldsymbol{f}_{i} \oplus \boldsymbol{f}_{fm} \oplus \boldsymbol{f}_{lm})$ using cross-modal attention (CA) with the key-value pairs ($\boldsymbol{\mathrm{K}_{ta}}, \boldsymbol{\mathrm{V}_{ta}}$) of merged text-audio embedding tokens $\boldsymbol{\mathrm{L}}(\boldsymbol{f}_{t} \oplus \boldsymbol{f}_{AS} \oplus \boldsymbol{f}_{AP})$ similar to the \eqref{eq:crossAttn1} as 
\begin{equation}\label{eq:crossAttn}
    \text{CA}(\boldsymbol{\mathrm{Q}_i}, \boldsymbol{\mathrm{K}_{ta}}, \boldsymbol{\mathrm{V}_{ta}}) = \text{SoftMax}\left(\frac{\boldsymbol{\mathrm{Q}_i}\boldsymbol{\mathrm{K}_{ta}}^\top}{\sqrt{d_k}}\right) \boldsymbol{\mathrm{V}_{ta}}.
\end{equation} 
As a result, the output latent embedding on audio-visual features $\boldsymbol{f}_{av}$ can serve as a compact and compressed representation of facial animation sequences in the high-dimensional space. Therefore, our next step is to learn a synthesizer \textit{i.e.,} a hierarchical latent diffusion model \cite{xu2024hallohierarchicalaudiodrivenvisual} for video generation and a corresponding MEL-spectrogram synthesizer based on the X-Text-to-Speech (XTTS) model \cite{casanova2024xttsmassivelymultilingualzeroshot}.  

\noindent\textbf{Latent Text Conditioned Spectrogram Synthesizer: }
The GPT-2 encoder is based on the TTS model \citep{casanova2023yourttszeroshotmultispeakertts} and \citep{shen2018naturalttssynthesisconditioning}. This part is composed of a decoder-only transformer module that is conditioned by the audio and speaker embedding vectors $\boldsymbol{f_{AS}}, \boldsymbol{f_{AP}}$ disentangled from the prompt-engineered audio embedding vector $\boldsymbol{f_{av}}$, and the auto-regressive generation of spectrogram tokens is fully driven by the input text tokens from $\boldsymbol{f_{av}}$. 
-

\noindent\textbf{Text-Anchored Audio-Video Latent Conditioned Denoising Diffusion: }
The Denoising Diffusion model aims to reverse a diffusion process\citep{ho2020denoisingdiffusionprobabilisticmodels,song2022denoisingdiffusionimplicitmodels} that progressively adds random Gaussian noise to data. Additionally, we employ an augmentation of the text-anchored latent embedding vector, learned to combine the audio and motion nuances within a single image, inside the Denoising U-Net \citep{ronneberger2015unetconvolutionalnetworksbiomedical}. %The model is initialized with pre-trained weights and fine-tuned during the training step. 

Throughout each step of the diffusion process, we introduce embedding cross-attention, which incorporates the combined latent space embedding, particularly our $\boldsymbol{f_{av}}$, into each diffusion step. This cross-attention mechanism allows the diffusion models to leverage the shared information across modalities, ensuring that the generated outputs (audio and video) are consistent with the input embedding. The inclusion of cross-attention helps to maintain coherence between the synthesised motion across all the pixels of the source image.

Additionally, diffusion cross-attention facilitates mutual information exchange between the audio and video diffusion blocks. This cross-attention mechanism enables the audio and video models to synchronise their outputs, ensuring that the generated audio and video components are temporally aligned. By integrating this cross-attention, our framework effectively coordinates the diffusion processes, leading to synchronised and coherent multimedia output.

\subsection{Decoding Phase for Audio-Video Generation}\label{subsec:DecoderSpace}
The outputs of the previous steps are processed by their respective final decoders. For audio generation, the synthesised spectrogram is passed through a Vocoder component of HiFi Generator module to obtain the final audio signal. For video, the Denoising UNet generates $f$ number of frames of dimension $\mathbb{R}^{4\times f \times 64 \times 64}$, which are decoded by a pre-trained decoder component of \citep{Kingma_2019} to produce the complete video.

\subsection{Loss Functions}\label{sec:LossFunctions} 
To train our model, we used \textbf{(1)} Video Loss as the Pixel-wise L1 Loss \textit{i.e.,} sum of the $N$ number of pixel intensities between the ground truth image frame $\mathcal{I}_{\text{gt}}^{f}$ and the generated frame $\mathcal{I}_{\text{gen}}^{f}$ for all the $f$ number of frames as $\mathcal{L_{\text{video}}} = \sum_f\sum_{i=1}^{N}\lVert (\mathcal{I}_{\text{gt}}^{f})^{i} - (\mathcal{I}_{\text{gen}}^{f})^i \rVert$, \textbf{(2)} Audio Loss as the Spectrogram MSE loss at the spectrogram $\mathcal{S}$ domain as mean squared error between the ground-truth magnitudes and generated magnitudes at different of time step $t$ as $\mathcal{S}_{\text{gt}}^{t}$ and the generated frame $\mathcal{S}_{\text{gen}}^{t}$ as $\mathcal{L_{\text{audio}}} = \frac{1}{T}\sum_{t\in T}\lVert (\mathcal{I}_{\text{gt}}^{f})^{i} - (\mathcal{I}_{\text{gen}}^{f})^i \rVert^2$. Total loss as $\mathcal{L_{\text{Total}}} 
= \lambda\mathcal{L_{\text{audio}}} + \mathcal{L_{\text{video}}}$ with regulations factor $\lambda$ = 0.1.

\section{Experimental Results}
\subsection{Datasets, Implementation and Evaluation}
\textbf{Datasets:} We have primarily conducted our experiments on 4 datasets. Our model training was done on a combination of \textbf{VoxCeleb} Dataset \citep{Nagrani19}, \textbf{FakeAVCeleb} dataset \citep{khalid2022fakeavcelebnovelaudiovideomultimodal}, \textbf{HDTF} \citep{zhang2021flow} and the \textbf{CelebV-HQ} dataset \citep{zhu2022celebvhqlargescalevideofacial}. For our experiment, we developed a subset of this dataset with 36000 videos. We chose this training set by filtering out individuals whose speech was in English from four datasets (VoxCeleb, FakeAVCeleb, CelebV-Hq and HDTF). The text transcribing was done using the Whisper model by OpenAI. For the source image, we just took the first frame of the video, and for the audio profile, we trimmed down the audio we extracted from the video to 2 seconds. We tested on more than 200 samples from each of the 4 datasets, resulting in a test set of over 800 unseen samples. 
%VoxCeleb is an audio-visual dataset consisting of short clips of human speech, extracted from interview videos uploaded to YouTube. FakeAVCeleb is a novel audio-video multimodal deepfake dataset. We only considered the non-deepfake part of the dataset. CelebV-HQ is a large-scale video facial attributes dataset demonstrating a diverse quality of data, which is important to test the robustness of our model. HDTF is a large in-the-wild high resolution audio-visual dataset built for talking face generation.

%We have taken approx 500 videos from each of these datasets. 500 - VoxCeleb 500 - CelebV 402 - FakeAVCeleb 409 - HDTF

\noindent\textbf{Preprocessing:} Our preprocessing involved resizing the videos to 512x512 and then cropping each video sample to the first 20 seconds (at 25FPS which equates to 500 frames). We then separated the audio from the video using ffmpeg, and then ran the OpenAI's Whisper model\citep{radford2022robustspeechrecognitionlargescale} to transcribe the audio speeches. %Our model can handle variable sequence lengths. To do so, we add the last 50 frames generated making it seamless. We ran inference on longer sequences and achieved consistent results.

\noindent\textbf{Implementation details:} The optimizer used for our model is AdamW with a learning rate of 1e-4 and weight decay of 1e-2, and the scheduler has a step-wise learning rate with a step size of 1000 and gamma of 0.5. The weight decay regularises the model, preventing any overfitting. We have used Nvidia 1xA6000s GPUs for training each model, and the model inference requires 12GB of VRAM. The total parameter size of the model comes to 1,575,936 and performs 5.39 GFLOPs (Giga Floating Point Operations) per generation.  We have trained the models for 10 epochs, with a batch size of 8. The Hifi-Gan, Wav2Vec Encoders, the Variational Autoencoder, Diffusion Models, and the GPT2 Decoder were pre-trained which were further trained with the rest of the entire proposed network.

\noindent\textbf{Evaluation Metrics:} Following are the evaluation matrices employed. For video generation, Fréchet Video Distance (FVD), Fréchet Inception Distance (FID), Fréchet Video Motion Distance(FVMD), Peak Signal-to-Noise Ratio (PSNR), Structural Similarity Index (SSIM), Kernel Video Distance (KVD), Learned Perceptual Image Patch Similarity (LPIPS) and Mean Opinion Score (MOS) were also employed.  For the MOS score, we involved 35 human experts, who scored the video and audio in range of 0-5.%\textit{Lip-Sync Error - Confidence}: Measures how well the lip movements align with the corresponding audio. Higher scores indicate better synchronization. %\textit{Lip-Sync Error - Distance}: Measures the average error between lip movements and audio signals. Lower distance indicates better synchronization. 
Metrics such as Fréchet Audio Distance (FAD), Short-Time Objective Intelligibility (STOI), Mel Cepstral Distortion(MCD), Word Error Rate (WER), Extended STOI (ESTOI), Perceptual Evaluation of Speech Quality (PESQ) and MOS were used. For AV synchronisation, Lip Sync Error Distance (LSE-D) and Lip Sync Error Confidence (LSE-C) were employed. %The higher the confidence, the better the audio-video correlation. %A lower confidence score denotes that there are several portions of the video with completely out-of-sync lip movements}.% From the below table we can conclude that our proposed model has performed better than the other SOTA models and is close to the ground truth, because we generate both the modalities.

\subsection{Result Analysis}
\noindent\textbf{Video Results}: From \autoref{tab:video_scores} and \autoref{vm}, we can observe that our model shows superior performance across all metrics employed on VoxCeleb, CelebV-Hq and HDTF. This indicates high fidelity and that minimal discrepancies are addressed by the proposed model. On the FakeAVCeleb, the performance is slightly poorer but can be comparable; it still maintains strong visual consistency and realism on visual inspection. For the CelebV-HQ, our model excels again, demonstrating its capability to produce high-quality video outputs. On HDTF, our model shows incredible performance in the FID and FVD metrics, beating all the other models, while our model is admirably performing considering FVMD when compared to Hallo. %Further, from , it can be observed that the proposed model has achieved better results than the state-of-the-art.

% First table (Video pipeline)
\begin{table}[t!]
\centering
\caption{Video pipeline evaluation scores across datasets.}
\label{tab:video_scores}
\small
\begin{tabular}{llccc}
\toprule
\textbf{Dataset} & \textbf{Model} & \textbf{FID ($\downarrow$)} & \textbf{FVD ($\downarrow$)} & \textbf{FVMD ($\downarrow$)} \\
\midrule
{\textbf{VoxCeleb}} & Audio2Head & 81.00 & 90.12 & 5100.92 \\
& Hallo & 67.28 & 70.69 & 5703.44 \\
& EAT & 85.16 & 80.38 & 4878.36 \\
& SadTalker & 119.36 & 112.77 & 6352.19 \\
& Our Model & \textbf{42.88} & \textbf{49.78} & \textbf{4192.07} \\
\midrule
{\textbf{FakeAVCeleb}} & Audio2Head & 93.59 & 97.85 & 1329.23 \\
& Hallo & \textbf{26.88} & \textbf{39.42} & 2351.20 \\
& EAT & 94.34 & 98.49 & \textbf{1324.91} \\
& SadTalker & 81.77 & 77.10 & 4158.18 \\
& Our Model & 47.24 & 49.15 & 2263.54 \\
\midrule
{\textbf{CelebV-HQ}} & Audio2Head & 90.22 & 102.76 & 2939.49 \\
& Hallo & 42.76 & 56.10 & 2816.68 \\
& EAT & 47.88 & 56.21 & 2894.31 \\
& SadTalker & 52.60 & 52.55 & 2789.19 \\
& Our Model & \textbf{34.01} & \textbf{43.67} & \textbf{2743.29} \\
\midrule
{\textbf{HDTF}} & Audio2Head & 37.78 & 32.69 & 2633.04 \\
& Hallo & 20.54 & 25.81 & \textbf{1290.57} \\
& EAT & 29.57 & 29.34 & 2573.05 \\
& SadTalker & 22.34 & 23.57 & 2410.89 \\
& Our Model & \textbf{11.72} & \textbf{15.58} & 1784.16 \\
\bottomrule
\end{tabular}
\end{table}

% Second table (Audio pipeline)

\begin{table}[t!]
\centering
\footnotesize
\setlength{\tabcolsep}{2.5pt}
\caption{Evaluation of video quality metrics.}
\label{tab:video_quality}
\small
\begin{tabular}{lccccc}
\toprule
\textbf{Method} & \textbf{PSNR ($\uparrow$)} & \textbf{SSIM ($\uparrow$)} & \textbf{KVD ($\downarrow$)} & \textbf{LPIPS ($\downarrow$)} & \textbf{MOS ($\uparrow$)} \\
\midrule
Hallo & 33.01 & 0.68 & 44.12 & 0.51 & 4.10 \\
% VASA & 33.83 & 0.65 & 43.12 & 0.49 & 4.09 \\
Audio2Head & 29.21 & 0.51 & 49.09 & \textbf{0.23} & 3.51 \\
EAT & 24.45 & 0.52 & 49.12 & 0.52 & 3.57 \\
SadTalker & 29.72 & 0.59 & 49.61 & 0.53 & 3.51 \\
Proposed & \textbf{35.94} & \textbf{0.73} & \textbf{41.22} & 0.54 & \textbf{4.22} \\
\bottomrule
\end{tabular}
\label{vm}
\end{table}

\begin{table}[t!]
\centering
\caption{Audio pipeline evaluation scores across datasets.}
\label{tab:audio_scores}
\small
\begin{tabular}{llccc}
\toprule
\textbf{Dataset} & \textbf{Model} & \textbf{FAD ($\downarrow$)} & \textbf{MCD ($\downarrow$)} & \textbf{STOI ($\uparrow$)} \\
\midrule
{\textbf{VoxCeleb}} & Tortoise & 258.54 & 82.37 & 0.10 \\
& Your\_TTS & \textbf{199.52} & 111.79 & \textbf{0.19} \\
& XTTS\_v2 & 249.17 & 100.80 & 0.13 \\
& GlowTTS & 329.21 & 103.94 & 0.15 \\
& Our Model & 241.75 & \textbf{75.39} & 0.17 \\
\midrule
{\textbf{FakeAVCeleb}} & Tortoise & 871.14 & 82.12 & 0.10 \\
& Your\_TTS & 445.38 & 65.60 & \textbf{0.21} \\
& XTTS\_v2 & 184.39 & 77.88 & 0.11 \\
& GlowTTS & 482.04 & 87.11 & 0.18 \\
& Our Model & \textbf{171.52} & \textbf{55.12} & 0.19 \\
\midrule
{\textbf{CelebV-HQ}} & Tortoise & 529.06 & 113.18 & 0.09 \\
& Your\_TTS & 520.01 & 137.58 & 0.16 \\
& XTTS\_v2 & 509.90 & 124.61 & 0.07 \\
& GlowTTS & 549.18 & 139.81 & \textbf{0.22} \\
& Our Model & \textbf{244.83} & \textbf{85.76} & 0.18 \\
\midrule
{\textbf{HDTF}} & Tortoise & 425.30 & 67.15 & 0.11 \\
& Your\_TTS & 467.42 & 49.38 & 0.15 \\
& XTTS\_v2 & 135.11 & 49.65 & 0.14 \\
& GlowTTS & 510.61 & 66.42 & 0.12 \\
& Our Model & \textbf{106.43} & \textbf{44.05} & \textbf{0.15} \\
\bottomrule
\end{tabular}
\end{table}

\begin{figure}[]
\centering
\includegraphics[width=7cm]{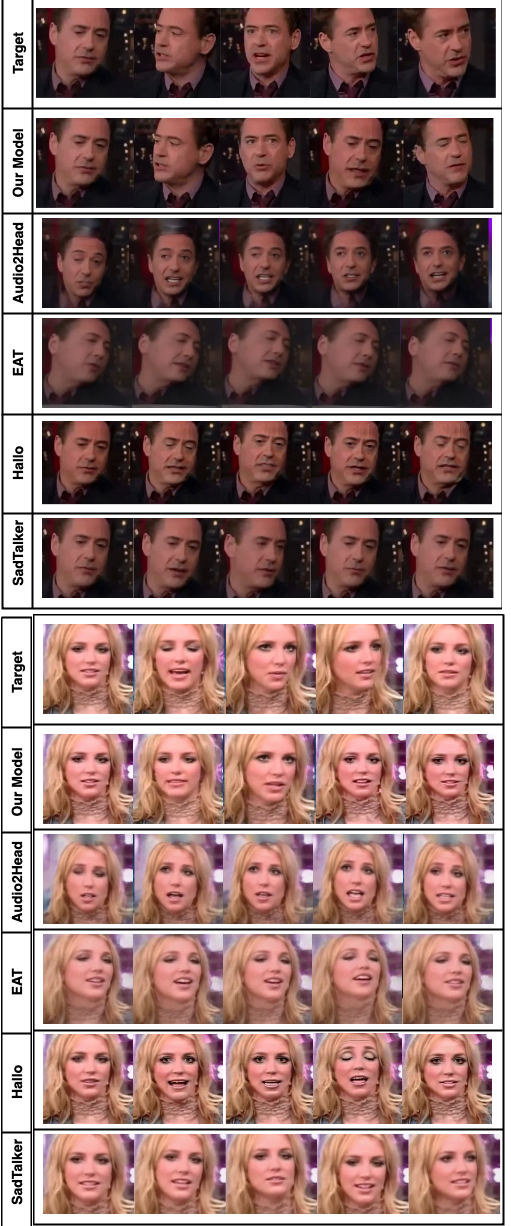}
\caption{Visual comparison on VoxCeleb, in the order: Ground Truth, Ours, Audio2Head, EAT, Hallo, and SadTalker. Columns represent 25s intervals.}
\label{fig}
\end{figure}
\setlength{\textfloatsep}{5pt plus 1.0pt minus 2.0pt}

\begin{figure}[]
\centering
\includegraphics[width=7cm]{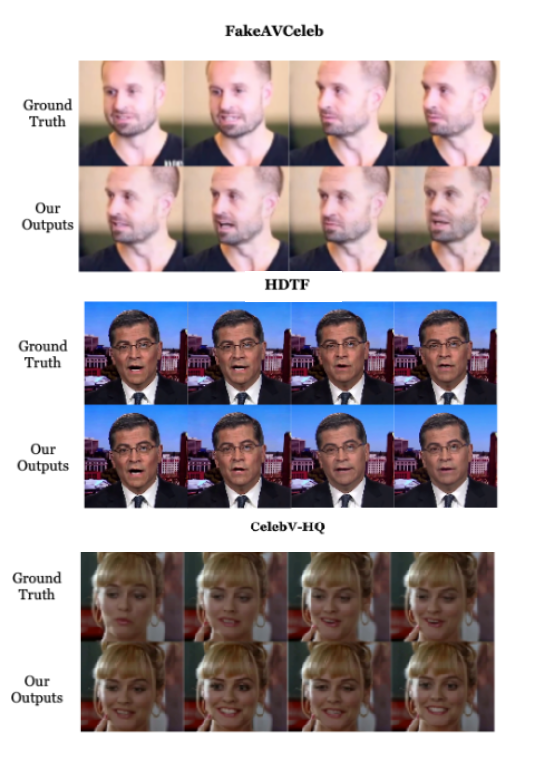}
\vspace{-10mm}
\caption{Results of our model on FakeAVCeleb, Celeb-HQ and HDTF datasets.}
\label{fig:sub3}
\end{figure}
\setlength{\textfloatsep}{5pt plus 1.0pt minus 2.0pt}

\begin{figure}[]
    \centering
    \begin{tabular}{cc}
        % Row 1 Images
        \includegraphics[width=0.2\textwidth]{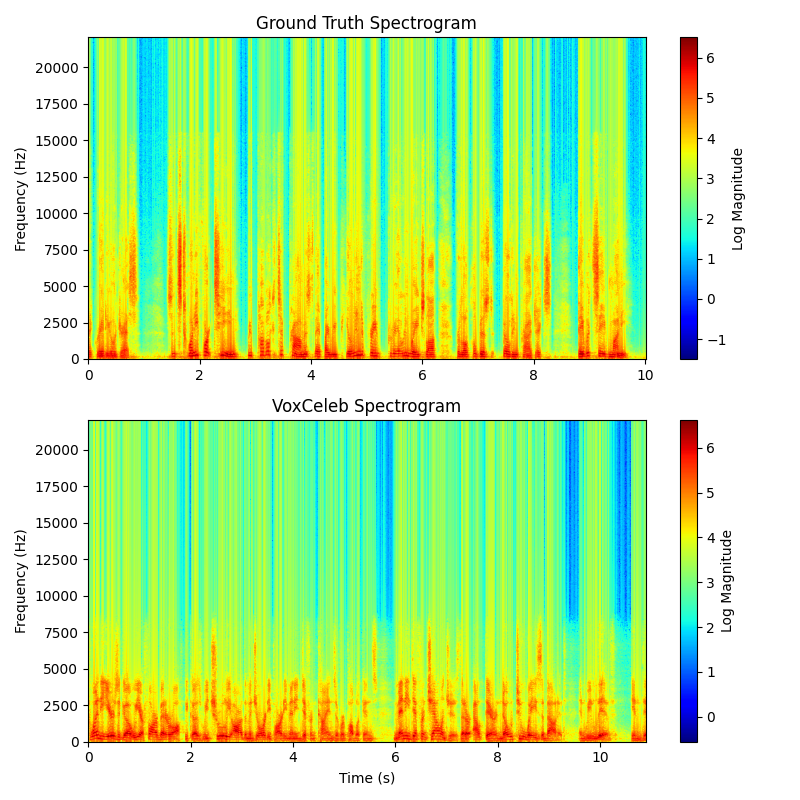} &
        \includegraphics[width=0.2\textwidth]{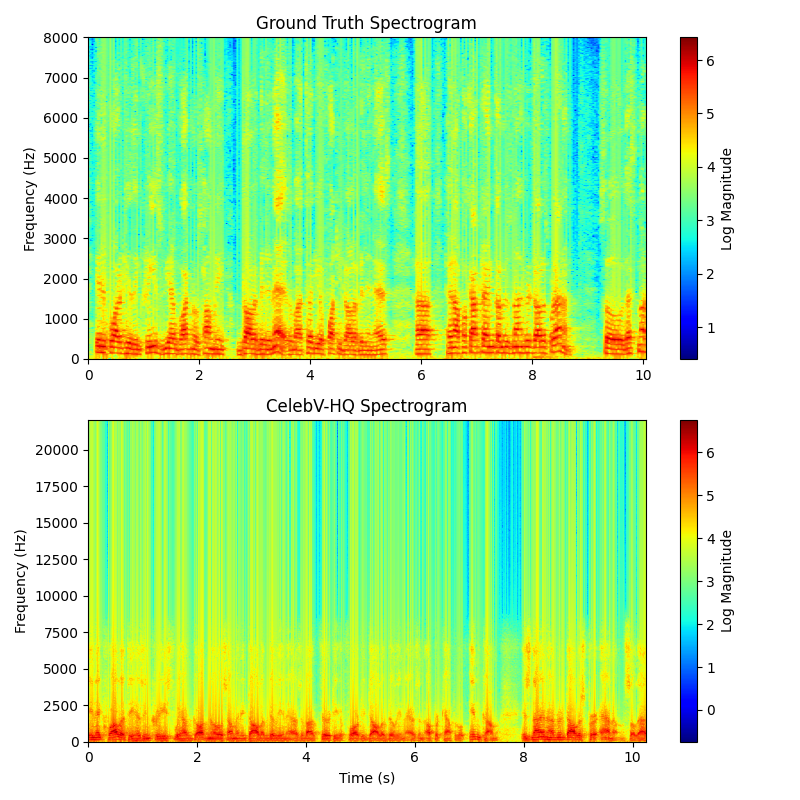} \\
        (a) & (b) \\
        
        % Row 2 Images
        \includegraphics[width=0.2\textwidth]{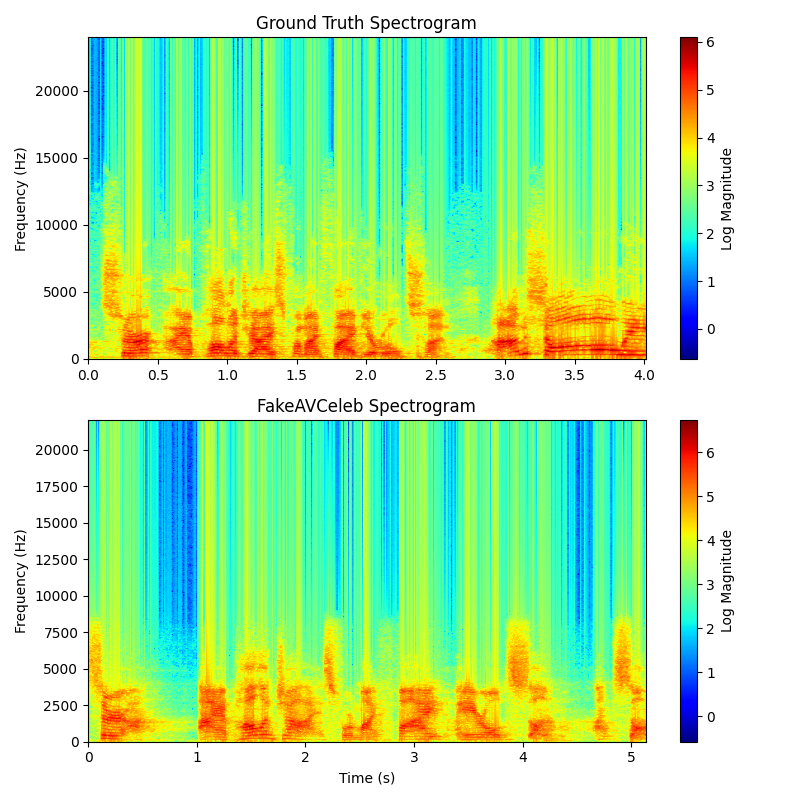} &
        \includegraphics[width=0.2\textwidth]{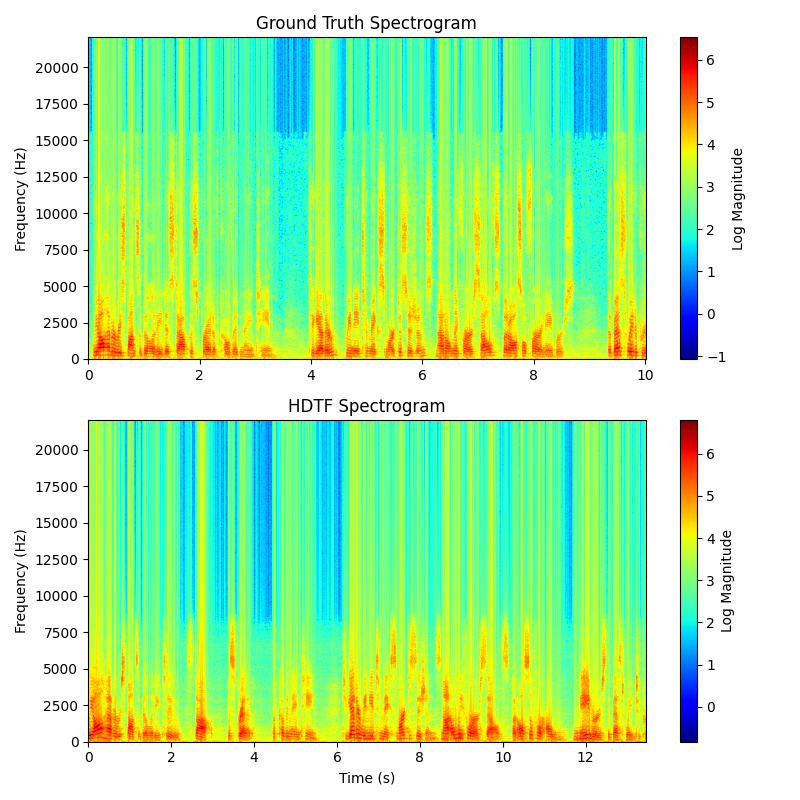} \\
        (c) & (d) \\
    \end{tabular}%
    \caption{Ground Truth vs. Generated Audio Spectrograms for (a) VoxCeleb, (b) CelebV-HQ, (c) FakeAVCeleb and (d) HDTF datasets}
    \label{fig:allfigs}
\end{figure}
\setlength{\textfloatsep}{5pt plus 1.0pt minus 2.0pt}

Based on the results, we observed that for some datasets, certain models work slightly better than the proposed model, and the reason behind this is that those models try to memorise certain properties from individual datasets. Whereas our model is a more generalized version that can be performed consistently on cross-datasets having varying resolution and video quality. The visualisation from Figure~\ref{fig} also concludes that our model can generate video very close to the ground truth and better than any model. From Figure~\ref{fig:sub3}  it can be concluded that our model can generate nearby results for HDTF, FakeAVCeleb and CelbV-HQ when compared to the ground truth.

\begin{table}[t!]
\centering
\caption{Evaluation of audio quality metrics.}
\label{tab:audio_quality}
\small
\begin{tabular}{lcccc}
\toprule
\textbf{Method} & \textbf{ESTOI ($\uparrow$)} & \textbf{PESQ ($\uparrow$)} & \textbf{WER ($\downarrow$)} & \textbf{MOS ($\uparrow$)} \\
\midrule
Tortoise & 0.42 & 3.80 & 0.21 & 4.32 \\
Your\_TTS & 0.31 & 3.71 & 0.30 & 3.61 \\
XTTS\_v2 & 0.33 & 3.51 & 0.34 & 3.21 \\
GlowTTS & 0.39 & 3.83 & 0.33 & 3.52 \\
Proposed & \textbf{0.43} & \textbf{4.01} & \textbf{0.18} & \textbf{4.56} \\
\bottomrule
\end{tabular}
\label{am}
\end{table}

% Third Table
\begin{table}[htbp]
\centering
\caption{Evaluation of audio-visual synchronisation.}
\label{tab:lip}
\begin{tabular}{lrr}
\toprule
\textbf{Model} & \textbf{LSE-C ($\uparrow$)} & \textbf{LSE-D ($\downarrow$)} \\
\midrule
Groundtruth & 5.45 & 8.52 \\
Hallo & 3.03 & 8.71 \\
Audio2Head & 2.51 & 10.34 \\
EAT & 4.39 & 9.35 \\
SadTalker & 5.44 & 10.09 \\
STE & 5.71 & 8.41 \\
\textbf{ETE/Proposed} & \textbf{5.74} & \textbf{8.38} \\
\bottomrule
\end{tabular}
\end{table}

\begin{table*}[t]
\centering
\caption{Ablation study of the transformer and cross-attention components. \textbf{STE}: Shared-TE, \textbf{ETE}: Explicit-TE, \textbf{DC}: Diffusion Cross-Attention, \textbf{EC}: Embedding Cross-Attention.}
\label{tab:ablation}
% Scale to 90% of the text width to make it "smaller"
\resizebox{0.8\textwidth}{!}{%
\begin{tabular}{cccccccccc}
\toprule
\multicolumn{4}{c}{\textbf{Components}} & \multicolumn{3}{c}{\textbf{Video Metrics}} & \multicolumn{3}{c}{\textbf{Audio Metrics}} \\
\cmidrule(r){1-4} \cmidrule(lr){5-7} \cmidrule(l){8-10}
\textbf{ETE} & \textbf{STE} & \textbf{DC} & \textbf{EC} & \textbf{FID ($\downarrow$)} & \textbf{FVD ($\downarrow$)} & \textbf{FVMD ($\downarrow$)} & \textbf{FAD ($\downarrow$)} & \textbf{MCD ($\downarrow$)} & \textbf{STOI ($\uparrow$)} \\
\midrule
& & \checkmark & \checkmark & 86.70 & 80.88 & 5275.89 & 328.27 & 95.44 & 0.07 \\
& \checkmark & & & 68.83 & 74.19 & 4412.74 & 260.91 & 87.51 & 0.11 \\
& \checkmark & \checkmark & & 63.68 & 71.38 & 4298.30 & 250.12 & 83.96 & 0.14 \\
& \checkmark & \checkmark & \checkmark & 61.44 & 69.15 & 2720.41 & 241.77 & 81.60 & 0.17 \\
\checkmark & & \checkmark & \checkmark & \textbf{42.88} & \textbf{49.78} & \textbf{4192.07} & \textbf{241.75} & \textbf{75.39} & \textbf{0.17} \\
\bottomrule
\end{tabular}
}
\end{table*}

\begin{table*}[t]
\centering
\caption{Ablation study of the encoder components.}
\label{tab:encoder_ablation}
\resizebox{0.8\textwidth}{!}{%
\begin{tabular}{lcccccc}
\toprule
\textbf{Ablation Setting} & \textbf{FID ($\downarrow$)} & \textbf{FVD ($\downarrow$)} & \textbf{FVMD ($\downarrow$)} & \textbf{FAD ($\downarrow$)} & \textbf{MCD ($\downarrow$)} & \textbf{STOI ($\uparrow$)} \\
\midrule
Only Visual Tokens Attended & 68.31 & 78.42 & 5747.04 & 304.98 & 81.17 & 0.13 \\
Only Audio Tokens Attended & 69.02 & 79.35 & 6576.85 & 301.49 & 80.65 & 0.13 \\
No audio sequence encoding & 85.25 & 94.28 & 7483.40 & 498.33 & 87.51 & 0.09 \\
No audio profile encoding & 70.10 & 80.96 & 5926.64 & 309.95 & 89.58 & 0.11 \\
No visual token encoding  & 54.38 & 62.02 & 5481.36 & \textbf{221.07} & \textbf{63.25} & 0.12 \\
\textbf{Proposed Model} & \textbf{42.88} & \textbf{49.78} & \textbf{4192.07} & 241.75 & 75.39 & \textbf{0.17} \\
\bottomrule
\end{tabular}
}
\end{table*}

\noindent\textbf{Audio Results:} We can infer from \autoref{tab:audio_scores} and \autoref{am} that our model consistently performs the best in considering all measures. The MCD Score metric suggests that it minimizes distortion between the spectral features of synthetic and reference speech. While considering the FAD scores, our model also performed on par with state-of-the-art, except on VoxCeleb where Your\_TTS is better; these showcase that the proposed model can generate consistently similar audio compared to the ground truth. Considering the STOI metric, the performance of our model is similar to or slightly lower than Your\_TTS. The analysis of all the measures showcases that our model is more generalized and realistic as it can minimise distortion and also generate accurate distributions, and maintain intelligibility of the speech consistently better than any other models. The visualization from Figure~\ref{fig:allfigs} also concludes that our model can generate audio very close to the ground truth.

\noindent\textbf{AV synchronisation results:} From Table~\ref{tab:lip}, we can conclude that our proposed model has performed better audio-video synchronization than SOTA and is close to the ground truth. The proposed model has the lowest LSE-D, i.e. better audio-visual match, i.e. and LSE-C i.e. better audio-video correlation. We have also analysed the model with varying accents, blurred audio profiles, and audio profiles of a kid with a source image of an adult and vice versa,  and the results were found to be very effective; no bias was found in any aspect. Models fail in a few scenarios where a very noisy audio profile is used, output audio is feeble or for source images with closed eyes, face dynamics get affected.

\subsection{Ablation Study}
\autoref{tab:ablation} shows the ablation study of our proposed model. We have 3 main sub-networks that define the output of our model. The \textbf{Transformer Encoder Block(TE)} \citep{vaswani2023attentionneed} with two variations shared-TE \textbf{(STE)}, where both audio and video pipeline share a transformer block and explicit-TE \textbf{(ETE)} where audio and video pipeline have explicit or separate transformer blocks. \textbf{Diffusion\citep{song2022denoisingdiffusionimplicitmodels} Cross Attention(DC)}, and the \textbf{Embedding Cross Attention(EC)}. From our results, it is understandable and explainable that the transformer encoder block, which encodes our inputs into a common latent space, is the most important modality of our network, with its removal drastically reducing our metric values. Our experiments also show that the cross-attention blocks between the diffusion models are more important than the embedding cross-attention, since our metric values drop more when we remove the diffusion cross-attention, probably since the diffusion cross-attention already syncs the modalities during the parallel learning stage. Another important aspect of ablation is the encoding latent in the individual transformer i.e. ETE is much better than STE. This implies that it is important to encode the latent for each modality separately while sharing information among the generated modalities. 

\autoref{tab:encoder_ablation} shows our ablation study on the encoders. ``Only Visual Tokens Attended" involves eliminating the audio prompt-guided transformer. Similarly, the ``Only Audio Tokens Attended" involves using only the audio prompt-guided transformer. ``No-audio sequence encoding" and ``No audio profile encoding" are results obtained by eliminating the encoding process of the Hifi-GAN and Wav2Vec models, respectively. ``No Visual token in prompt-guided-Transformer" involves not attending to the visual tokens in the prompt-guided-Transformer. 

\vspace{-3mm}
\subsection{Social Risks and Mitigations}
There are social risks with technology development for text-driven audio-video talking face generation in regards to privacy, malicious acts, etc. To mitigate such risk, ethical guidance for the use of such generation techniques is required. By addressing these, we aim to promote responsible and ethical generative technology.

\section{Conclusion}

This paper introduces a novel method for realistic speaking and talking faces by joint multimodal video and audio generation. We provide a holistic architecture where the information is exchanged between the modalities via the proposed multi-entangled latent space. A source image of an individual as a driving frame, reference audio, which can be referred to as the audio profile of the individual and a driving or prompt text is passed as an input. The model encodes the input driving image, prompt/driving text, and the voice profile, which are further combined and passed to the proposed multi-entangled latent space consisting of two separate transformers and a diffusion block for video and text decoder for audio pipeline to foster key-value and query representation for each modality. By this spatiotemporal person-specific features between the modalities are also established. The entangled-based learning representation is further passed to the respective decoder of audio and video modality
for respective outputs. Conducted experiments and ablation studies prove that the proposed multi-entangled latent-based learning representation has helped our model obtain superior results on both video and audio outputs as compared to state-of-the-art models. While there is always scope for improvement in the future, we believe that our model has shown a promising new learning representation for realistic speaking and talking face generation models. 
\vspace{-1mm}

\section*{Acknowledgements}

This work was funded by the  BITS Pilani-funded New Faculty Seed Grant (NFSG) under the project titled ``Playing God-generation of virtual environment through masked auto-encoders and diffusion technique” and project number NFSG/HYD/2023/H0831.
% {
%     \small
%     \bibliographystyle{ieeenat_fullname}
%     \bibliography{main}
% }

\end{document}